\def\year{2015}
\documentclass[letterpaper]{article}
\usepackage{aaai16}
\usepackage{times}
\usepackage{helvet}
\usepackage{courier}
\usepackage{graphicx} 
\usepackage{subfig}
\usepackage{natbib}

\usepackage{algorithm}
\usepackage{algorithmic}
\usepackage{amsmath}
\usepackage{url}
\usepackage{amsmath}
\usepackage{amssymb}
\usepackage{mathrsfs}
\usepackage{array}
\usepackage{bm}
\usepackage{multirow}
\usepackage{hyperref}

\def\E{{\rm E}}

\def\I{{\bf I}}
\def\tI{\tilde{\bf I}}
\def\F{{\bf F}}

\def\obs{{\rm obs}}
\def\syn{{\rm syn}}

\renewcommand{\algorithmicrequire}{\textbf{Input:}}
\renewcommand{\algorithmicensure}{\textbf{Output:}}

\begin{document}
%
\title{Learning FRAME Models Using CNN Filters}
\author{Yang Lu, Song-Chun Zhu, Ying Nian Wu\\
	Department of Statistics, University of California, Los Angeles, USA
}
\maketitle
\begin{abstract}
\begin{quote}
The convolutional neural network (ConvNet or CNN) has proven to be very successful in many tasks such as those  in computer vision. In this conceptual paper, we study the generative perspective of the discriminative CNN. In particular, we propose to learn the generative FRAME (Filters, Random field, And Maximum Entropy) model using the highly expressive filters pre-learned by the CNN at the convolutional layers. We show that the learning algorithm can generate  realistic and rich object and texture patterns in natural scenes. We explain that each learned model corresponds to a new CNN unit at a layer above the layer of filters employed by the model. We further show that it is possible to learn a new layer of CNN units using a generative CNN model, which is a product of experts model, and the learning algorithm admits an EM interpretation with binary latent variables. 
\end{quote}
\end{abstract}

\section{Introduction}

The breakthrough made by the convolutional neural network (ConvNet or CNN)  \citep{krizhevsky2012imagenet, lecun1998gradient} on the ImageNet dataset \citep{deng2009imagenet} was a watershed event that has transformed the fields of computer vision and speech recognition as well as related industries. While CNN has proven to be a powerful discriminative machine, researchers have recently become increasingly interested in the generative perspective of CNN. An interesting example is the recent work of Google deep dream (http://deepdreamgenerator.com/). Although it did not smash any performance records, it did capture people's imagination by generating interestingly vivid images. 

In this conceptual paper, we explore the generative perspective of CNN more formally by defining generative models based on CNN features, and learning these models by generating images  from the models. Adopting the metaphor of Google deep dream, we let the generative models dream by generating images. But unlike the google deep dream, we learn the models from real images by making the dreams come true. 

 Specifically, we propose to learn the FRAME (Filters, Random field, And Maximum Entropy) models \citep{zhu1997minimax, xie2014learning} using the highly  nonlinear filters pre-learned by CNN at the convolutional layers. A FRAME model is a random field model that defines a probability distribution on the image space. The model is generative in the sense that images can be generated from the probability distribution defined by the  model. The probability distribution is the maximum entropy distribution that reproduces the statistical properties of filter responses in the observed images. Being of the maximum entropy, the distribution is the most random distribution that matches the observed statistical properties of filter responses, so that images sampled from this distribution can be considered typical images that share the statistical properties of the observed images. 
 
There are two versions of FRAME models in the literature. The original version is a stationary model developed for modeling texture patterns \citep{zhu1997minimax}. The more recent version is a non-stationary extension designed to represent object patterns  \citep{xie2014learning}. Both versions of the FRAME models can be sparsified by selecting a subset of filters from a given dictionary. 

The filters used in the FRAME model are the oriented and elongated Gabor filters at different scales, as well as the isotropic Difference of Gaussian (DoG) filters of different sizes. These are linear filters that capture simple local image features such as edges and blobs. With the emergence of the more expressive nonlinear filters learned by CNN at various convolutional layers, it is only natural to replace the linear filters in the original FRAME models by the CNN filters in the hope of learning more expressive models. 

 We use the Langevin dynamics to sample from the probability distribution defined by the model.  Such a dynamics was first applied to the FRAME model  by \cite{zhu1997GRADE}, and the gradient descent part of the dynamics was interpreted as the Gibbs Reaction And Diffusion Equations (GRADE). When applied to the FRAME model with CNN filters, the dynamics can be viewed as a recurrent generative form of the model, where the reactions and diffusions are governed by the CNN filters of positive and negative weights respectively. 
    
Incorporating CNN filters into the FRAME model is not an ad hoc utilitarian exploit. It is actually a seamless meshing between the FRAME model and the CNN model. The original FRAME model has an energy function that consists of a layer of linear filtering followed by a layer of pointwise nonlinear transformation. It is natural to follow the deep learning philosophy to add alternative layers of linear filtering and nonlinear  transformation to have a deep FRAME model that directly corresponds to a CNN. More importantly, the learned  FRAME model using CNN filters corresponds to a new CNN unit at the layer directly above the layer of CNN filters employed by the FRAME model. In particular, the non-stationary FRAME becomes a single CNN node at a specific position where the object appears, whereas the stationary FRAME becomes a special type of convolutional unit. Therefore, the learned  FRAME model can be viewed as a generative version of CNN unit. 

In addition to learning a single CNN unit, we can also learn a  new layer of multiple convolutional units from non-aligned images, so that each convolutional unit represents one type of local pattern. We call the resulting model the generative CNN model. It is a product of experts model \citep{Hinton2002a}, where each expert models a mixture of activation and inactivation of a local pattern. The rectified linear unit can be justified as an approximation to the energy function of this mixture model. The learning algorithm admits an interpretation in terms of the EM algorithm \citep{dempster1977maximum} with a hard-decision E-step that  detects the local patterns modeled by the convolutional units. 

The main purpose of this paper is to establish the conceptual correspondence between the generative FRAME model and the discriminative CNN, thus providing a  formal generative perspective for CNN. Such a perspective is much needed because it may eventually lead to unsupervised learning of CNN in a generative fashion without the need for image labeling. 

\section{Past work}

Recently there have been many interesting papers on visualizing CNN nodes, such as deconvolutional networks \citep{zeiler2013visualizing}, score maximization \citep{SimonyanCoRR2013}, and the recent artful work of Google deep dream (http://deepdreamgenerator.com/)  and painting style \citep{German2015a}. Our work is different from these previous methods in that we learn a rigorously defined generative model from training images, and the learned models correspond to new CNN units. This work is a continuation of the recent work on generative CNN \citep{Dai2015ICLR}. 

There have also been recent papers on generative models based on supervised image generation \citep{Alexey2015}, variational auto-encoders \citep{Hinton95thewake-sleep, KingmaCoRR13, RezendeICML2014, MnihGregor2014, Kulkarni2015, KarolICML2015}, and adversarial networks \citep{Denton2015a}. Each of these papers  learns a top-down multi-layer model for image generation, but the parameters of the top-down generation model are completely separated from the parameters of the bottom-up recognition model.  Our work seeks to learn a generative model based on the knowledge learned by the bottom-up recognition model, i.e., the image generation model and the image recognition model share the same set of weight parameters.

\section{FRAME models based on linear filters} 

This section reviews the background on the FRAME models based on linear filters. 

Let $\I$ be an image defined on a square (or rectangular) domain ${\cal D}$.   Let $\{F_k, k = 1, ..., K\}$ be a bank of linear filters, such as elongate and oriented Gabor filters at different scales, as well as isotropic Difference of Gaussian (DoG) filters of different sizes. Let $F_k *\I$ be the filtered image or feature map, and $[F_k*\I](x)$ be the filter response at position $x$ ($x$ is a two-dimensional coordinate). A linear filter $F_k$ can be written as a two-dimensional function $F_k( x)$, so that $[F_k *\I](y) = F_k( x) \I(y+x)$, which is a translation invariant linear operation. 

The original FRAME model \citep{zhu1997minimax}  for texture patterns is a stationary or spatially homogeneous Markov random field or Gibbs distribution of the following form: 
\begin{eqnarray}
   p(\I; \lambda) = \frac{1}{Z(\lambda)} \exp \left[\sum_{k=1}^{K} \sum_{x \in {\cal D}} \lambda_{k} \left([F_k*\I](x)\right)\right], 
   \label{eq:FRAME}
\end{eqnarray}
where $\lambda_k()$ is a nonlinear function to be estimated from the training images, $\lambda = (\lambda_k(), k = 1, ..., K)$, and $Z(\lambda)$ is the normalizing constant to make $p(\I; \lambda)$ integrate to 1. In the original paper of \cite{zhu1997minimax}, each $\lambda_k()$ is discretized and estimated as a step function, i.e., $\lambda_k(r) = \sum_{b=1}^{B} w_{k, b} h_b(r)$, where $b \in \{1, ..., B\}$ indexes the equally spaced bins of discretization, and $h_b(r) = 1$ if $r$ is in bin $b$, and 0 otherwise, i.e., $h() = (h_b(), b = 1, ..., B)$ is a 1-hot indicator vector, and $\sum_x h([F_k*\I](x))$ is the marginal histogram of filter map $F_k*\I$. The spatially pooled marginal histograms are the sufficient statistics of model (\ref{eq:FRAME}). 

Model (\ref{eq:FRAME})  is stationary because the function $\lambda_k()$ does not depend on position $x$. This stationary model is used to model texture patterns. In model (\ref{eq:FRAME}),  the energy function $U(\I; \lambda) = - \sum_{k} \sum_{x} \lambda_{k}( [F_k*\I](x))$ involves a layer of linear filtering by $\{F_k\}$, followed by a layer of pointwise nonlinear transformation by $\{\lambda_k()\}$. Repeating this pattern recursively (while also adding local max pooling and sub-sampling) will lead to a generative version of CNN. 

The  non-stationary or spatially inhomogeneous FRAME model for object patterns  \citep{xie2014learning} is  of the following form: 
\begin{eqnarray}
   p(\I; \lambda) = \frac{1}{Z(\lambda)} \exp \left[\sum_{k = 1}^{K} \sum_{x \in {\cal D}} \lambda_{k, x} ([F_k*\I](x))\right] q(\I) ,
   \label{eq:iFRAME}
\end{eqnarray}
where the function $\lambda_{k, x}()$ depends on position $x$, and $\lambda = (\lambda_{k, x}(), \forall k, x)$.  Again $Z(\lambda)$ is the normalizing constant.  The model is non-stationary because $\lambda_{k, x}()$ depends on position $x$. It is impractical to estimate $\lambda_{k, x}()$ as a step function at each $x$, so $\lambda_{k, x}()$ is parametrized as a one-parameter function 
\begin{eqnarray}
\lambda_{k, x}(r) = w_{k, x} h(r), 
\label{eq:iFRAME2}
\end{eqnarray}
 where $h()$ is a pre-specified rectification function, and $w = (w_{k, x}, \forall k, x)$ are the unknown parameters to be estimated. In the paper of \cite{xie2014learning} , they use $h(r) = |r|$ for full wave rectification. One can also use rectified linear unit $h(r) = \max(0, r)$   \citep{krizhevsky2012imagenet} for half wave rectification, which can be considered an elaborate two-bin discretization. $q(\I)$ is a reference distribution, such as the Gaussian white noise model
 \begin{eqnarray}
    q(\I) = \frac{1}{(2\pi\sigma^2)^{|{\cal D}|/2}} \exp\left[- \frac{1}{2\sigma^2} ||\I||^2\right], 
    \label{eq:Gaussian}
\end{eqnarray}   
where $|{\cal D}|$ counts the number of pixels in the image domain ${\cal D}$. 

 In the original FRAME model (\ref{eq:FRAME}), $q(\I)$ is assumed to be a uniform measure.  In model (\ref{eq:iFRAME}), we can also absorb $q(\I)$, in particular, the $ \frac{1}{2\sigma^2} ||\I||^2$ term,  into the energy function, so that the model is again defined relative to a uniform measure as in the original FRAME model (\ref{eq:FRAME}). We make $q(\I)$ explicit here because we shall specify the parameter $\sigma^2$  instead of learning it, and use $q(\I)$ as the null model for the background.  In models (\ref{eq:iFRAME}) and (\ref{eq:iFRAME2}), $(w_{k, x}, \forall x, k)$ can be considered a second-layer linear filter on top of the first layer filters $\{F_k\}$ rectified by $h()$. 
 
Both models (\ref{eq:FRAME}) and (\ref{eq:iFRAME}) can be sparsified. Model (\ref{eq:FRAME}) can be sparsified by selecting a small set of filters $F_k$ using the filter pursuit procedure \citep{zhu1997minimax}.  Model (\ref{eq:iFRAME}) can be sparsified by selecting a small number of filters $F_k$ and positions $x$, so that only a small number of $w_{k, x}$ are non-zero. The sparsification can be achieved by a shared matching pursuit method \citep{xie2014learning} or a generative boosting method \citep{xie2015boosting}. 

\section{FRAME models based on CNN filters}

Instead of using linear filters, we can use the filters at various convolutional layers of a pre-learned CNN. Suppose there exists a bank of filters $\{F_k, k = 1, ..., K\}$ (e.g., $K = 512$) at a certain convolutional layer of a pre-learned CNN. For an image $\I$ defined on the square image domain ${\cal D}$, let $F_k*\I$ be the feature map of filter $F_k$, and let $[F_k*\I](x)$ be the filter response of $\I$ to $F_k$ at position $x$ (again $x$ is a two-dimensional coordinate).  We assume that $[F_k*\I](x)$ is the response obtained after applying the rectified linear transformation $h(r) = \max(0, r)$. Then the non-stationary FRAME model becomes
\begin{eqnarray}
   p(\I; w) = \frac{1}{Z(w)} \exp \left[\sum_{k=1}^{K} \sum_{x \in {\cal D}} w_{k, x} [F_k*\I](x)\right] q(\I), 
   \label{eq:CNN-FRAME}
\end{eqnarray}
where $q(\I)$ is again the Gaussian white noise model (\ref{eq:Gaussian}) , and $w  = (w_{k, x}, \forall k, x)$ are the unknown parameters to be learned from the training data.  $Z(w)$ is the normalizing constant. Model (\ref{eq:CNN-FRAME}) shares the same form as model (\ref{eq:iFRAME}) with linear filters, except that the rectification function $h()$ in model (\ref{eq:iFRAME}) is already absorbed in the CNN filers $\{F_k\}$ in model (\ref{eq:CNN-FRAME}) with $h(r) = \max(0, r)$.  We shall use  model (\ref{eq:CNN-FRAME}) for generating object patterns. 

The stationary  FRAME model is of the following form:
\begin{eqnarray}
   p(\I; w) = \frac{1}{Z(w)} \exp \left[\sum_{k=1}^{K} \sum_{x \in {\cal D}} w_{k} [F_k*\I](x)\right] q(\I), 
   \label{eq:sFRAME}
\end{eqnarray}
which is almost the same as model (\ref{eq:CNN-FRAME}) except that  $w_k$ is the same across $x$. $w = (w_k, \forall k)$. We shall use model (\ref{eq:sFRAME}) for generating texture patterns. 

Again, both models (\ref{eq:CNN-FRAME}) and (\ref{eq:sFRAME}) can be sparsified, either by forward selection such as filter pursuit \citep{zhu1997minimax} or generative boosting  \citep{xie2015boosting}, or by backward elimination.

\section{Learning and sampling algorithm} 

The basic learning algorithm for object model estimates the unknown parameters $w$ from a set of aligned training images $\{\I_m, m = 1, ..., M\}$ that come from the same object category,  where $M$ is the total number of training images.  In the basic learning algorithm, the weight parameters $w$ can be estimated by maximizing the  log-likelihood function
\begin{equation}
L(w) = \frac{1}{M} \sum_{m=1}^{M} \log p(\I_m; w), \label{eq:logl0}
\end{equation}
where $p(\I; w)$ is defined by (\ref{eq:CNN-FRAME}). $L(w)$ is a concave function. The first derivatives of $L(w)$ are 
\begin{equation}
\frac{\partial L(w)}{\partial w_{k, x}} = \frac{1}{M} \sum_{m=1}^{M} [F_k*\I_m](x) - \E_{w}\left([F_{k}*\I](x) \right), 
\label{eq:gradient}
\end{equation}
where $\E_{w}$ denotes the expectation with respect to $p(\I; w)$. The expectation can be approximated by Monte Carlo integration. The second derivative of $L(w)$ is the variance-covariance matrix of $([F_k*\I](x), \forall k, x)$. 
 $w$ can be computed by a stochastic gradient ascent algorithm \citep{younes1999convergence}:
\begin{equation}
\begin{aligned}
&w^{(t+1)}_{k, x} = w^{(t)}_{k, x} + 
\\\ & \gamma \Bigg[\frac{1}{M} \sum_{m=1}^{M} [F_k *\I_m](x)
               - \frac{1}{\tilde{M}} \sum_{m=1}^{\tilde{M}} [F_k *\tI_m](x)\Bigg]
\end{aligned}
\label{eq:learning}
\end{equation}   
for every $k \in \{1, ..., K\}$ and $x \in {\cal D}$, where $\gamma$ is the learning rate, and $\{\tI_m\}$ are the synthesized images sampled from $p(\I; w^{(t)})$ using MCMC. $\tilde{M}$ is the total number of independent parallel Markov chains that sample from $p(\I; w^{(t)})$.  The learning rate $\gamma$ can be made inversely proportional to the observed variance of $\{[F_k*\I_m](x), \forall m\}$, as well as being inversely proportional to the iteration $t$ as in stochastic approximation. 

In order to sample from $p(\I; w)$, we adopt the Langevin dynamics. Writing the energy function 
\begin{eqnarray}
   U(\I, w) = - \sum_{k=1}^{K} \sum_{x \in {\cal D}} w_{k, x} [F_k*\I](x) +  \frac{1}{2\sigma^2} ||\I||^2.
   \label{eq:energy}
\end{eqnarray}
The Langevin dynamics iterates 
\begin{eqnarray}
     \I_{\tau+1} = \I_{\tau} - \frac{\epsilon^2}{2}  U'(\I_\tau, w) + \epsilon Z_\tau, 
     \label{eq:Langevin}
\end{eqnarray}
where $U'(\I, w) = \partial U(\I, w)/\partial \I$. This gradient can be computed by back-propagation. In (\ref{eq:Langevin}), $\epsilon$ is a small step-size, and $Z_\tau \sim {\rm N}(0, {\bf 1})$, independently across $\tau$, where the bold font ${\bf 1}$ is the identify matrix, i.e., $Z_\tau$ is a Gaussian white noise image whose pixel values follow ${\rm N}(0, 1)$ independently. Here we use $\tau$ to denote the time steps of the Langevin sampling process, because $t$ is used for the time steps of the learning process. The Langevin sampling process is an inner loop within the learning process. Between every two consecutive updates of $w$ in the learning process,  we run a finite number of iterations of the Langevin dynamics starting from the images generated by the previous iteration of the learning algorithm, a scheme called ``warm start'' in the literature. The Langevin equation was also adopted by \cite{zhu1997GRADE}, who called the corresponding gradient descent algorithm the Gibbs reaction and diffusion equations (GRADE).  

\begin{algorithm}
\caption{Learning and sampling algorithm}
\label{code:FRAME}
\begin{algorithmic}[1]

\REQUIRE ~~\\
(1)  training images $\{\I_m, m=1,...,M\}$ \\
(2) a filter bank $\{F_{k},  k = 1, ..., K\}$\\
(3) number of synthesized images $\tilde{M}$\\
(4) number of Langevin steps $L$\\
(5) number of learning iterations $T$

\ENSURE~~\\
(1) estimated parameters $w=(w_{k, x}, \forall k, x)$\\
(2) synthesized images $\{\tI_m, m = 1, ..., \tilde{M}\}$ 

\item[]
\STATE Calculate observed statistics:

$H^{\obs}_{k, x} \leftarrow \frac{1}{M}\sum_{m=1}^{M} [F_k*\I_m](x), \forall k, x.$ 
\STATE Let $t\leftarrow 0$, initialize $w^{(0)}_{k, x} \leftarrow 0, \forall k, x$.
\STATE Initialize $\tI_m \leftarrow 0$, for $m = 1, ..., \tilde{M}$. 
\REPEAT 
\STATE For each $m$, run $L$ steps of Langevin dynamics to update $\tI_m$, i.e., starting from the current $\tI_m$, each step updates
$ \tI_{m} \leftarrow \tI_{m} - \frac{\epsilon^2}{2} U'(\tI_m, w^{(t)}) + \epsilon Z$,  
where $Z \sim {\rm N}(0, {\bf 1})$. 
\STATE Calculate synthesized statistics:\\
$H^{\syn}_{k, x} \leftarrow \frac{1}{\tilde{M}}\sum_{m=1}^{\tilde{M}} [F_k*\tI_m](x), \forall k, x.$
\STATE Update $w^{(t+1)}_{k, x} \leftarrow w^{(t)}_{k, x}+ \gamma ( H^{\obs}_{k, x} - H^{\syn}_{k, x}) $, $\forall k, x$.

\STATE Let $t \leftarrow t+1$
\UNTIL $t = T$
\end{algorithmic}
\end{algorithm}

Algorithm \ref{code:FRAME} describes the details of the learning and sampling algorithm.  Algorithm \ref{code:FRAME} embodies the principle of ``analysis by synthesis,''  i.e., we generate synthesized images from the current model, and then update the model parameters based on the difference between the synthesized images and the observed images. If we consider the synthesized images as ``dreams'' of the current model (following the metaphor used by Google deep dream), then the learning algorithm is to make the dreams come true. 

From the MCMC perspective, Algorithm \ref{code:FRAME} runs non-stationary parallel Markov chains that sample from a Gibbs distribution with a changing energy landscape,  like in  simulated annealing or tempering. This may help the chains to avoid the trapping of local modes.  We can also use ``cold start'' scheme by initializing Langevin dynamics from white noise images in each learning iteration and allowing the dynamics enough time to relax. 

For learning stationary FRAME (\ref{eq:sFRAME}), usually $M = 1$, i.e., we observe one texture image, and we update the parameters
\begin{equation}
\begin{aligned}
&w^{(t+1)}_{k} = w^{(t)}_{k} +
 \\ & \frac{ \gamma}{|{\cal D}|} \Bigg[\frac{1}{M} \sum_{m=1}^{M} \sum_{x \in {\cal D}} [F_k *\I_m](x)
               - \frac{1}{\tilde{M}} \sum_{m=1}^{\tilde{M}} \sum_{x \in {\cal D}} [F_k *\tI_m](x)\Bigg]
\end{aligned}
\label{eq:learning1}
\end{equation}   
for every $k \in \{1, ..., K\}$, where there is a spatial pooling across positions $x \in {\cal D}$. The sampling is again accomplished by Langevin dynamics. The learning and sampling algorithm for the stationary model (\ref{eq:sFRAME}) only involves minor modifications of Algorithm \ref{code:FRAME}. 

\begin{figure}
	\centering
	\setlength{\fboxrule}{1pt}
	\setlength{\fboxsep}{0cm}
	\subfloat{
		\includegraphics[width=.99\linewidth]{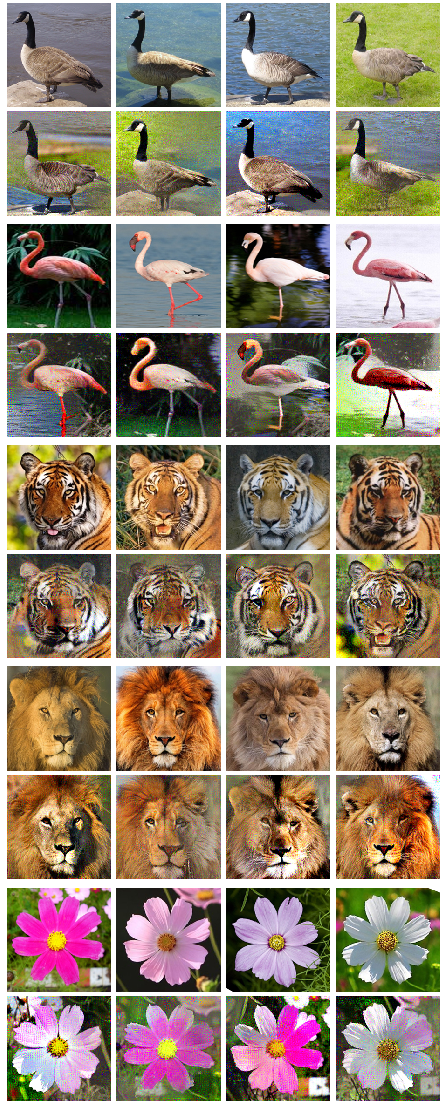}
	}
	\caption{Generating object patterns. For each category, the first row displays 4 of the training images, and the second row displays generated images.}
	\label{fig:object}
\end{figure}
   
\section{Generative CNN units} 

On top of the convolutional layer of filters $\{F_k, k = 1, ..., K\}$, we can build another layer of filters $\{\F_j, j = 1, ..., J\}$ (with $\F$ in bold font, and indexed by $j$), so that 
\begin{eqnarray}
[\F_j*\I](y) = h\left( \sum_{k,  x} w^{(j)}_{k,   x}  [F_{k} * \I](y+x) +b_j\right),
\label{eq:filter}
\end{eqnarray}
where $h()$ is a rectification function such as the rectified linear unit $h(r) = \max(0, r)$, and where the bias term $b_j$ is related  to $-\log Z(w)$. For simplicity, we ignore the layers of local max pooling and sub-sampling.  

Model (\ref{eq:CNN-FRAME}) corresponds to a single filter in $\{\F_j\}$ at a particular position $y$ (e.g.,  the origin $y = 0$) where we assume that the object appears. The weights $(w^{(j)}_{k, x})$ can be learned by fitting model (\ref{eq:CNN-FRAME}) using Algorithm \ref{code:FRAME}, which enables us to add a CNN node in a generative fashion. 

The log-likelihood ratio of the object model $p(\I; w)$ in (\ref{eq:CNN-FRAME}) versus the  background model $q(\I)$ is $\log (p(\I; w)/q(\I))= \sum_{k} \sum_{x} w_{k, x} [F_k*\I](x) - \log Z(w)$. It  can be used as a score for detecting the object versus the background. If the score is below a threshold, no object is detected, and the score is rectified to 0.  The rectified linear unit $h()$ in $\F_j$ in (\ref{eq:filter}) accounts for the fact that at any position $y$, the object either appears or not. More formally, consider a mixture model $p(\I) = \alpha p(\I; w) + (1-\alpha) q(\I)$,  where $\alpha$ is the frequency that the object is activated, and $1-\alpha$ is the frequency of background. $\log (p(\I)/q(\I)) = \log (1+ \exp (\sum_{k} \sum_{x} w_{k, x} (F_k*\I)(x) - \log Z(w) + \log (\alpha/(1-\alpha)) + \log(1-\alpha)$. We can approximate the soft max function $\log (1+e^r)$ by the hard max function $\max(0, r)$. Thus we can identify the bias term as $b = \log(\alpha/(1-\alpha)) - \log Z(w)$, and the rectified linear unit models a mixture of ``on'' and ``off'' of an object pattern.

\begin{figure}
	\centering
	\setlength{\fboxrule}{1pt}
	\setlength{\fboxsep}{0cm}	

\subfloat{
	\includegraphics[width=.99\linewidth]{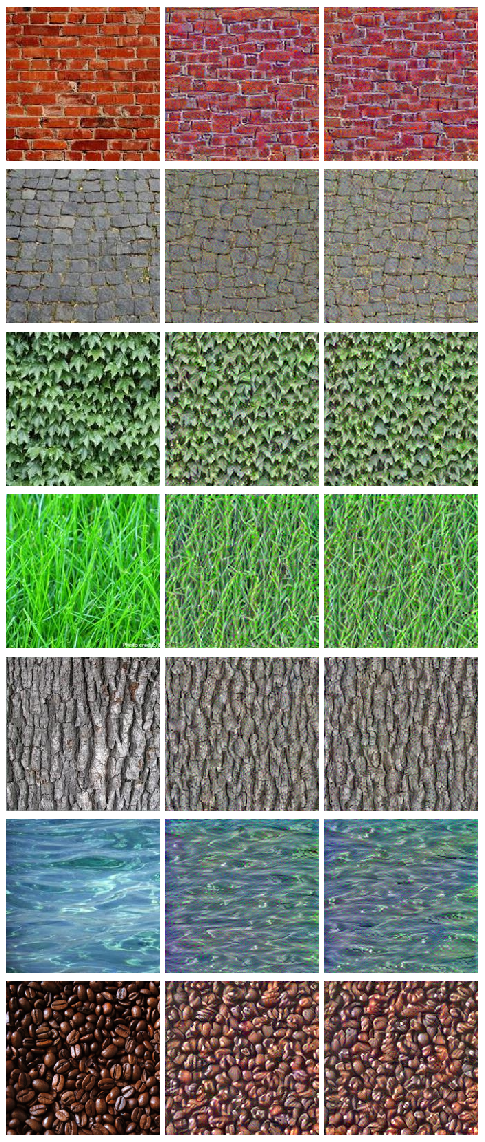}
}
	\caption{Generating texture patterns. For each category, the first image is the training image, and the next 2 images are generated images. }
	\label{fig:texture}
\end{figure}

Model (\ref{eq:CNN-FRAME}) is used to model images where the objects are aligned and are from the same category. For non-aligned images that may consist of multiple local patterns, we can extend model (\ref{eq:CNN-FRAME}) to a convolutional version with multiple filters
\begin{eqnarray}
   p(\I; w) = \frac{1}{Z(w)} \exp \left[\sum_{j=1}^{J} \sum_{x \in {\cal D}} [\F_j*\I](x)\right] q(\I), 
   \label{eq:CNN-FRAME1}
\end{eqnarray}
where $\{\F_j\}$ are defined by (\ref{eq:filter}). This model is a product of experts model \citep{Hinton2002a}, where each  $[\F_j*\I](x)$ is an expert about a mixture of an activation or inactivation of a local pattern of type $j$ at position $x$.  We call model (\ref{eq:CNN-FRAME1}) with (\ref{eq:filter}) the generative CNN model. The model can also be considered a dense version of the And-Or model \citep{ZhuM06}, where the binary switch of each expert corresponds to an Or-node, and the product corresponds to an And-node.

\begin{figure}
	\centering
	\setlength{\fboxrule}{1pt}
	\setlength{\fboxsep}{0cm}
	
%
\subfloat{
	\includegraphics[width=.99\linewidth]{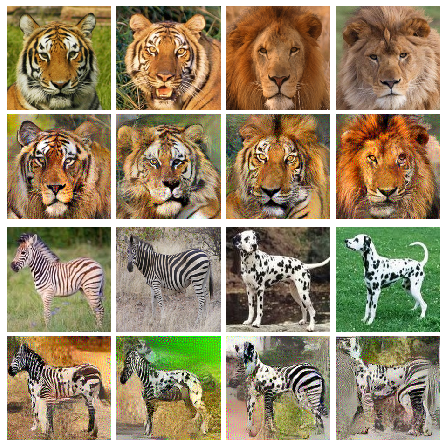}
}
	\caption{Generating hybrid object patterns.  For each experiment, 
		the first row displays 4 of the training images, and the second row displays generated images.  }
	\label{fig:hybrid1}
\end{figure}

The stationary model (\ref{eq:sFRAME}) corresponds to a special case of  generative CNN model (\ref{eq:CNN-FRAME1}) with (\ref{eq:filter}), where there is only one $j$, and $[\F*\I](x) = \sum_{k=1}^{K} w_k [F_k*\I](x)$, which is a special case of (\ref{eq:filter}) without rectification. It is a singleton filter that combines lower layer filter responses at the same position. 

More importantly, due to the recursive nature of CNN, if the weight parameters $w_k$ of the stationary model  (\ref{eq:sFRAME}) are absorbed into the filters $F_k$ by multiplying the weight and bias parameters of each $F_k$ by $w_k$,  then the stationary model becomes the generative CNN model (\ref{eq:CNN-FRAME1}) except that the top-layer filters $\{\F_j\}$ are replaced by the lower layer filters $\{F_k\}$.  The  learning of  the stationary model  (\ref{eq:sFRAME}) is a simplified version of the learning of the generative CNN model  (\ref{eq:CNN-FRAME1})  where there is only one multiplicative parameter $w_k$  for each filter $F_k$. The learning of  the stationary model  (\ref{eq:sFRAME}) is more unsupervised and more indicative of the expressiveness of the CNN features than the learning of the non-stationary model (\ref{eq:CNN-FRAME}) because the former does not require alignment.

Suppose we observe $\{\I_m, m = 1, ..., M\}$ from  the generative CNN model (\ref{eq:CNN-FRAME1}) with (\ref{eq:filter}). Let $L(w) = \frac{1}{M} \sum_{m=1}^{M} \log p(\I_m; w)$ be the log-likelihood where $p(\I; w)$ is defined by (\ref{eq:CNN-FRAME1}) and (\ref{eq:filter}), then 
\begin{equation}
\begin{aligned}
\frac{\partial L(w)}{\partial w_{k, x}^{(j)}} =& \frac{1}{M} \sum_{m=1}^{M}\sum_{y \in {\cal D}} \delta_{j, y}(\I_m)  [F_{k} * \I_m](y+x) \\
      &   -  \E_{w} \left[ \sum_{y \in {\cal D}} \delta_{j, y}(\I)  [F_{k} * \I](y+x)\right], 
\end{aligned}
\label{eq:generativeGradient}
\end{equation}
where 
\begin{eqnarray}
\delta_{j, y}(\I) = h'\left( \sum_{k,  x} w^{(j)}_{k,   x}  [F_{k} * \I](y+x) +b_j\right)
\label{eq:detector}
\end{eqnarray}
is a binary on/off detector of the local pattern of type $j$ at position $y$ on image $\I$,  because for $h(r) = \max(0, r)$, $h'(r) = 0$ if $r  \leq  0$, and $h'(r) = 1$ if $r > 0$.  The gradient (\ref{eq:generativeGradient}) admits an EM \citep{dempster1977maximum} interpretation which is typical in unsupervised learning algorithms that involve latent  variables. Specifically, $\delta_{j, y}()$ detects the local pattern of type $j$  modeled by $\F_j$. This step can be considered a hard-decision E-step. With the local patterns detected,  the parameters of $\F_j$ are then updated in a similar way as in (\ref{eq:learning}),  which can be considered the M-step. That is, we learn $\F_j$ only from image patches where we detect pattern  $j$. Such a scheme was used by \cite{Hongyi2014} to learn codebooks of active basis models \citep{AB}.

\begin{figure}
	\centering
	\setlength{\fboxrule}{1pt}
	\setlength{\fboxsep}{0cm}
\subfloat{
	\includegraphics[width=.99\linewidth]{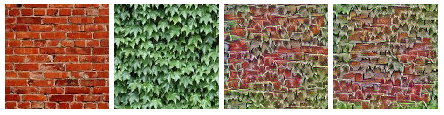}
}
			\caption{Generating hybrid texture patterns. 
		The first 2 images are training images, and the last 2 images are generated images. }
	\label{fig:hybrid2}
\end{figure}

Model (\ref{eq:CNN-FRAME1}) with (\ref{eq:filter}) defines a recursive scheme, where the learning of higher layer filters $\{\F_j\}$ is based on the lower layer filters $\{F_k\}$.  We can use this recursive scheme to build up the layers from scratch. We can start from the ground  layer of the raw image, and learn the first layer filters. Then based on the first layer filters, we learn the second layer filters, and so on.  

After building up the model layer by layer, we can continue to refine the parameters of all the layers simultaneously. In fact, the parameter $w$ in model (\ref{eq:CNN-FRAME1}) can be interpreted more broadly as multi-layer connection weights that define all the layers of filters. The gradient of the log-likelihood is 
\begin{equation}
\begin{aligned}
\frac{\partial L(w)}{\partial w} =& \frac{1}{M} \sum_{m=1}^{M}  \sum_{j=1}^{J} \sum_{x \in {\cal D}} \frac{\partial}{\partial w} [\F_j*\I_m](x) \\
      &   -  \E_{w} \left[  \sum_{j=1}^{J} \ \sum_{x \in {\cal D}} \frac{\partial}{\partial w} [\F_j*\I](x)\right], 
\end{aligned}
\label{eq:generativeGradient1}
\end{equation}
where $\partial [\F_j*\I](x)/\partial w$ involves multiple layers of binary detectors. The resulting algorithm also requires partial derivative $\partial [\F_j *\I](x)/\partial \I$ for Langevin sampling, which can be considered a recurrent generative model driven by the binary switches at multiple layers.  Both $\partial [\F_j *\I](x)/\partial w$ and $\partial [\F_j *\I](x)/\partial \I$  are readily available via back-propagation. See \cite{hinton2006unsupervised, Ng2011} for earlier work along this direction.  See also   \cite{Dai2015ICLR} for generative gradient of CNN.

\begin{figure}
	\centering
	\setlength{\fboxrule}{1pt}
	\setlength{\fboxsep}{0cm}
		\subfloat{
			\includegraphics[width=.99\linewidth]{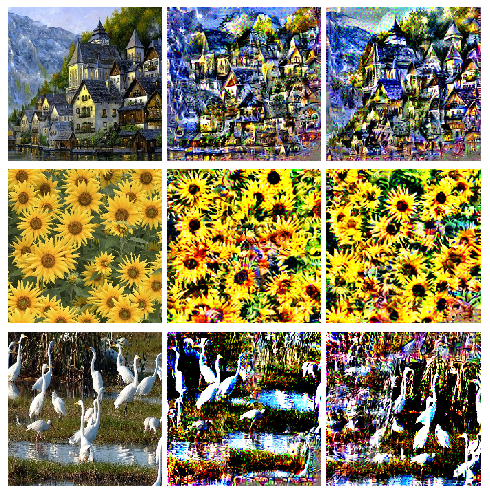}
		}
		\caption{Learning without alignment.  
		In each row, the first  image is the training image, and the next  2  images are generated images. }
	\label{fig:nonalign1}
\end{figure}

Finally, we can also learn a FRAME model based on the features at the top fully connected layer, 
\begin{eqnarray}
   p(\I; W) =  \frac{1}{Z(W)}  \exp \left[  \sum_{i=1}^{N} W_i [\Psi_i*\I] \right]q(\I), 
   \label{eq:fullyFRAME}
\end{eqnarray}
where $\Psi_i$ is the $i$-th feature at the top fully connected layer,  $N$ is the total number of features at this layer (e.g., $N = 4096$), and $W_i$ are the parameters, $W = (W_i, \forall i)$.  $\Psi_i$ can still be viewed as a filter whose filter map is 1 $\times$ 1. Suppose there are a number of image categories, and suppose we learn a model (\ref{eq:fullyFRAME}) for each image category with a category-specific $W$. Also suppose we are given the prior frequency of each category. A simple exercise of the Bayes rule then gives us the soft-max classification rule for the posterior probability of category given image, which is   the discriminative CNN. 

\section{Image generation experiments}

In our experiments, we use VGG filters \citep{simonyan2014very}, and we use the Matlab code of MatConvNet \citep{matconvnn}.

\begin{figure}
	\centering
	\setlength{\fboxrule}{1pt}
	\setlength{\fboxsep}{0cm}	
	\subfloat{
		\includegraphics[width=.99\linewidth]{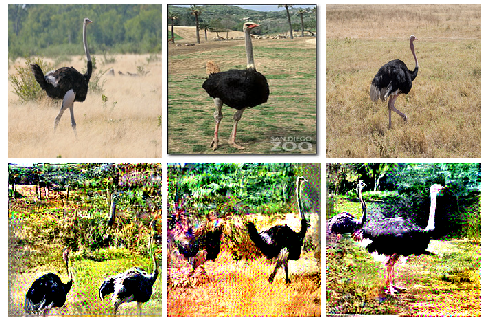}
	}
		\caption{Learning from non-aligned images.  
		The first row displays 3 of the training images, and the second row displays 3 generated images.  }
	\label{fig:nonalign2}
\end{figure}

{\bf Experiment 1: generating object patterns.} We  learn the non-stationary FRAME model (\ref{eq:CNN-FRAME}) from  images of aligned objects. The images are collected from the internet. For each category, the number of training images is around 10. We use $\tilde{M} = 16$  parallel chains for Langevin sampling. The number of Langevin iterations between every two consecutive updates of the parameters is $L = 100$.  Fig. \ref{fig:object} shows some experiments using filters from the 3rd convolutional layer of VGG.  For each experiment, the first row displays 4 of the training images, and the second row displays 4 of the synthesized images generated by Algorithm \ref{code:FRAME}.

{\bf Experiment 2: generating texture patterns.} We  learn the stationary FRAME model (\ref{eq:sFRAME}) from images of textures. Fig. \ref{fig:texture} shows some experiments. Each experiment is displayed in one row, where the first image is the training image, and the other two images are generated by the learning algorithm. 

{\bf Experiment 3: generating hybrid patterns}. We  learn models (\ref{eq:CNN-FRAME}) and  (\ref{eq:sFRAME}) from images of mixed categories, and generate hybrid patterns. Figs. \ref{fig:hybrid1} and \ref{fig:hybrid2} display a few examples.

{\bf Experiment 4: learning a new layer of filters from non-aligned images}.  We learn the generative CNN model (\ref{eq:CNN-FRAME1}) with (\ref{eq:filter}). Fig. \ref{fig:nonalign1} displays 3 experiments. In each row, the first image is the training image, and the next 2 images are generated by the learned model. In the first scenery experiment, we learn 10 filters at the 4th convolutional layer (without local max pooling), based on the pre-trained VGG filters at the 3rd convolutional layer. The size of each Conv4 filter to be learned is $11 \times 11 \times 256$. In the sunflower and egret experiments, we learn 20 filters of size $7 \times 7 \times 256$ (with local max pooling).  Clearly these learned filters capture the local patterns and re-shuffle them seamlessly. Fig. \ref{fig:nonalign2} displays an experiment where we learn a layer of filters from a small training set of non-aligned images. The first row displays 3 examples of training images and the second row displays the generated images. We use the same parameter setting as in the sunflower experiment. These experiments show that it is possible to learn generative CNN model (\ref{eq:CNN-FRAME1}) from non-aligned images.

\section{Conclusion}

In this paper, we learn the FRAME models based on pre-trained CNN filters. It is possible to learn the multi-layer FRAME model or the generative CNN model (\ref{eq:CNN-FRAME1})  from scratch in a layer by layer fashion without relying on pre-trained CNN filters. The learning will be a recursion of model (\ref{eq:CNN-FRAME1}) , and it can be unsupervised without image labeling.

\section*{Code and data}

The code, data, and more experimental results can be found at {\url{http://www.stat.ucla.edu/~yang.lu/project/deepFrame/main.html} }

\section*{Acknowlegements}

The code in our work is based on the Matlab code of MatConvNet \citep{matconvnn}, and our experiments are based on the VGG features \citep{simonyan2014very}. We are grateful to these authors for sharing their code and results with the community. 

We thank Jifeng Dai for earlier collaboration on generative CNN. We thank Junhua Mao and Zhuowen Tu for sharing their expertise on CNN.  We thank one reviewer for deeply insightful comments. The work is supported by NSF DMS 1310391, ONR MURI N00014-10-1-0933, DARPA SIMPLEX  N66001-15-C-4035,  and DARPA MSEE   FA 8650-11-1-7149.

\section*{Appendix 1.  Maximum entropy justification}

The FRAME model (\ref{eq:CNN-FRAME})  can be justified by the maximum entropy or minimum divergence principle. Suppose the true distribution that generates the observed images $\{\I_m\}$ is $f(\I)$. Let $w^{\star}$ solve  the population version of the maximum likelihood equation: 
\begin{eqnarray}
\E_{w}([F_k*\I](x)) = \E_{f}([F_k*\I](x)), \;\forall k, x. 
\end{eqnarray}
 Let $\Omega$ be the set of all the feasible distributions $p$ that share the statistical properties of $f$ as captured by $\{F_k\}$:
 \begin{eqnarray}
\Omega = \{p:  \E_{p}([F_k*\I](x)) = \E_{f}([F_k*\I](x)) \; \forall k, x\}. 
 \end{eqnarray}
Then it can be shown that among all $p \in \Omega$, $p(\I; w^{\star})$ achieves the minimum of ${\rm KL}(p||q)$, i.e., the Kullback-Leibler divergence from $p$ to $q$ \citep{della1997inducing}. Thus $p(\I; w^{\star})$ can be considered the projection of $q$ onto $\Omega$, or the minimal modification of the reference distribution $q$ to match the statistical properties of the true distribution $f$.  In the special case where $q$ is a uniform distribution, $p(\I; w^{\star})$ achieves the maximum entropy among all distributions in $\Omega$. For Gaussian white noise $q$, as mentioned before, we can absorb the $\frac{\|\I\|^2}{2\sigma^2}$ term into the energy function as in (\ref{eq:energy}), so model (\ref{eq:CNN-FRAME}) can be written relative to a uniform measure with $\|\I\|^2$ as an additional feature. The maximum entropy interpretation thus still holds if we opt to estimate $\sigma^2$ from the data. 
 
 \section*{Appendix 2.  Julesz ensemble justification}

 The learning algorithm seeks to match statistics of the synthesized images to those of the observed images, as indicated by (\ref{eq:learning}) and (\ref{eq:learning1}), where the difference between the observed statistics and the synthesized statistics drives the update of the parameters. If the algorithm converges, and if the number of the synthesized images $\tilde{M}$ is large in the case of object patterns or if the image domain ${\cal D}$ is large in the case of  texture patterns, then the synthesized statistics should match the observed statistics. Assume $q(\I)$ to be the uniform distribution for now. We can consider the following ensemble in the case of object patterns: 
 \begin{equation}
 \begin{aligned}
     {\cal J} = \bigg\{(\tilde{\I}_m, & m = 1, ..., \tilde{M}):                
    \frac{1}{\tilde{M}} \sum_{m=1}^{\tilde{M}} [F_k *\tI_m](x)\\
    & =  \frac{1}{M} \sum_{m=1}^{M} [F_k *\I_m](x), \forall k, x \bigg\}. 
  \end{aligned}
  \label{eq:Julesz1}
 \end{equation}     
Consider the uniform distribution over ${\cal J}$. Then as $\tilde{M} \rightarrow \infty$, the marginal distribution of any $\tilde{\I}_m$ is given by model (\ref{eq:CNN-FRAME}) with $w$ being estimated by maximum likelihood.  Conversely, model (\ref{eq:CNN-FRAME})  puts uniform distribution on ${\cal J}$ if $\tilde{\I}_m$ are independent samples from model (\ref{eq:CNN-FRAME}) and if $\tilde{M} \rightarrow \infty$. 

As for the texture model, we can take $\tilde{M} = 1$, but let the image size go to $\infty$. First fix the square domain ${\cal D}$. Then embed it at the center of a larger square domain $\overline{\cal D}$. Consider the ensemble of images defined on $\overline{\cal D}$: 
 \begin{equation}
 \begin{aligned}
     {\cal J} = \bigg\{\tilde{\I}:                
   &\frac{1}{|\overline{\cal D}|}\sum_{x \in \overline{\cal D}} [F_k *\tI](x) \\&=  \frac{1}{|{\cal D}|} \frac{1}{M} \sum_{m=1}^{M} \sum_{x \in {\cal D}}  [F_k *\I_m](x), \forall k \bigg\}. 
  \end{aligned}
  \label{eq:Julesz2}
 \end{equation}     
Then under the uniform distribution on ${\cal J}$, as $|\overline{\cal D}| \rightarrow \infty$, the distribution of $\tilde{\I}$ restricted to ${\cal D}$ is given by model (\ref{eq:sFRAME}). Conversely, model (\ref{eq:sFRAME}) defined on $\overline{\cal D}$ puts uniform distribution on ${\cal J}$ as $|\overline{\cal D}| \rightarrow \infty$. 

The ensemble ${\cal J}$ is called the Julesz ensemble by \cite{wu2000texture}, because Julesz  was the first to pose the question as to what statistics define a texture pattern \citep{Julesz1962}. The averaging across images in equation (\ref{eq:Julesz1}) enables re-mixing of  the parts of the observed images to generate new object images.  The spatial averaging in equation (\ref{eq:Julesz2}) enables  re-shuffling of the local patterns in the observed image to generate a new texture image. That is, the averaging operations lead to exchangeability. 

For object patterns, define the discrepancy 
\begin{eqnarray}
\Delta_{k, x} =  \frac{1}{\tilde{M}} \sum_{m=1}^{\tilde{M}} [F_k *\tI_m](x) - \frac{1}{M} \sum_{m=1}^{M} [F_k *\I_m](x).
\end{eqnarray}
One can sample from the uniform distribution on ${\cal J}$ in (\ref{eq:Julesz1}) by running a simulated annealing algorithm that samples from $p(\tilde{\I}_m, m = 1, ..., \tilde{M}) \propto \exp(-\sum_{k, x} \Delta_{k, x}^2/T)$ by Langevin dynamics while gradually lowering the temperature $T$, or simply by gradient descent as in \cite{German2015a} by assuming $T = 0$. The sampling algorithm is very similar to Algorithm \ref{code:FRAME}.  One can use a similar method to sample from the uniform distribution over ${\cal J}$ in (\ref{eq:Julesz2}).  Such a scheme was used by \cite{zhu2000texture} for texture synthesis. 

In the above discussion, we assume $q(\I)$ to be the uniform distribution. If $q(\I)$ is Gaussian, we only need to add the feature $\|\I\|^2$ to the pool of features to be matched. The above results still hold. 

The Julesz ensemble perspective connects statistics matching and the FRAME models, thus providing another justification for these models in addition to the maximum entropy principle.

\bibliographystyle{aaai}
\bibliography{deepframe}
  
\end{document}